\documentclass[10pt,twocolumn,letterpaper]{article}

\usepackage{wacv}
\usepackage{times}
\usepackage{epsfig}
\usepackage{graphicx}
\usepackage{amsmath}
\usepackage{amssymb}
\usepackage{fixltx2e}
\usepackage[font=footnotesize,justification=centering]{caption}
\usepackage[pagebackref=true,breaklinks=true,letterpaper=true,colorlinks,bookmarks=false]{hyperref}
\usepackage[pagebackref=true,breaklinks=true,letterpaper=true,colorlinks,bookmarks=false]{hyperref}

\wacvfinalcopy % *** Uncomment this line for the final submission

\def\wacvPaperID{470} % *** Enter the wacv Paper ID here

\ifwacvfinal\fi
\begin{document}

%%%%%%%%% TITLE
\title{Visual Recognition of Paper Analytical Device Images for Detection of Falsified Pharmaceuticals}

% Authors at the same institution
%\author{First Author \hspace{2cm} Second Author \\
% Authors at different institutions
\author{Sandipan Banerjee\textsuperscript{1}, James Sweet\textsuperscript{1}, Christopher Sweet\textsuperscript{1}, and Marya Lieberman\textsuperscript{2}\\
\textsuperscript{1}Department of Computer Science and Engineering,\\
\textsuperscript{2}Department of Chemistry and Biochemistry,\\
University of Notre Dame\\
{\tt\small \{sbanerj1, jsweet, csweet1, mlieberm\}@nd.edu}
}

\maketitle
\ifwacvfinal\thispagestyle{empty}\fi

%%%%%%%%% ABSTRACT
\begin{abstract}
%Counterfeiting drugs is a big issue in a majority of the developing countries, especially in Africa. We have developed a set of inexpensive paper cards, called Paper Analytical Devices (PADs), which can efficiently classify drugs based on their chemical composition, as a potential solution to the problem. These cards have different reagents embedded in them which produce different color tones after reacting with the chemical compounds in different drug solutions. Each drug-reagent pair produces a unique set of color descriptors. The same descriptors are not obtained for a counterfeit version of the same drug; this difference is perceivable by humans. However, diffusion of the reactants into the vertical lanes creates lot of noise, which often hampers the recognition process. To deal with this, we propose an automatic visual recognition system for use with these PAD cards. At first, the optimal set of reagents is found by running singular value decomposition on the intensity values of the color blobs in the cards. A dataset of cards embedded with these reagents is produced to generate the most unique results for a set of 26 drugs. In the second stage, we train two popular convolutional neural network (CNN) models, with the card images. We also extract some ``hand-designed" features from the images and train a nearest neighbor classifier and a non-linear support vector machine with them. On testing all the methods, the CNN models reached an accuracy of over 94\% and outperformed the hand-designed classifiers (methods).

Falsification of medicines is a big problem in many developing countries, where technological infrastructure is inadequate to detect these harmful products. We have developed a set of inexpensive paper cards, called Paper Analytical Devices (PADs), which can efficiently classify drugs based on their chemical composition, as a potential solution to the problem. These cards have different reagents embedded in them which produce a set of distinctive color descriptors upon reacting with the chemical compounds that constitute pharmaceutical dosage forms. If a falsified version of the medicine lacks the active ingredient or includes substitute fillers, the difference in color is perceivable by humans. However, reading the cards with accuracy takes training and practice, which may hamper their scaling and implementation in low resource settings. To deal with this, we have developed an automatic visual recognition system to read the results from the PAD images. At first, the optimal set of reagents was found by running singular value decomposition on the intensity values of the color tones in the card images. A dataset of cards embedded with these reagents is produced to generate the most distinctive results for a set of 26 different active pharmaceutical ingredients (APIs) and excipients. Then, we train two popular convolutional neural network (CNN) models, with the card images. We also extract some ``hand-crafted" features from the images and train a nearest neighbor classifier and a non-linear support vector machine with them. On testing, higher-level features performed much better in accurately classifying the PAD images, with the CNN models reaching the highest average accuracy of over 94\%.

\end{abstract}
%%%%%%%%% BODY TEXT

\section{Introduction}
\begin{figure}[t]
\begin{center}
   \includegraphics[width=1.0\linewidth]{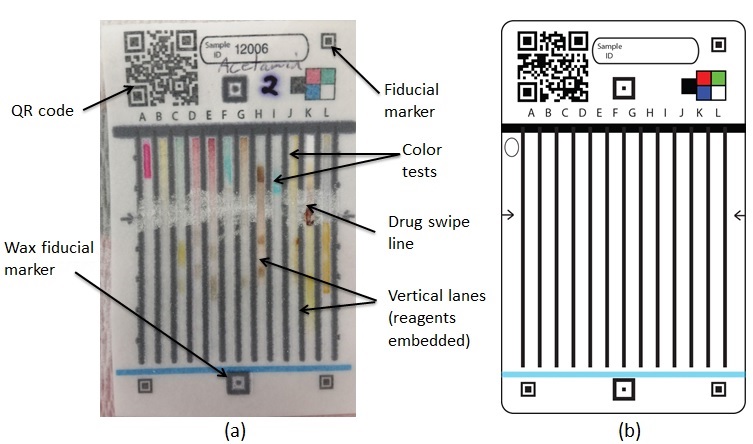}
\end{center}
   \caption{(a) Unrectified PAD image of the drug \emph{acetaminophen} with different reagents in the 12 different vertical lanes, (b) PAD image artwork.}
\label{fig:PADExamples}
%\label{fig:onecol}
\end{figure}
\begin{figure}[t]
\begin{center}
   \includegraphics[width=0.9\linewidth]{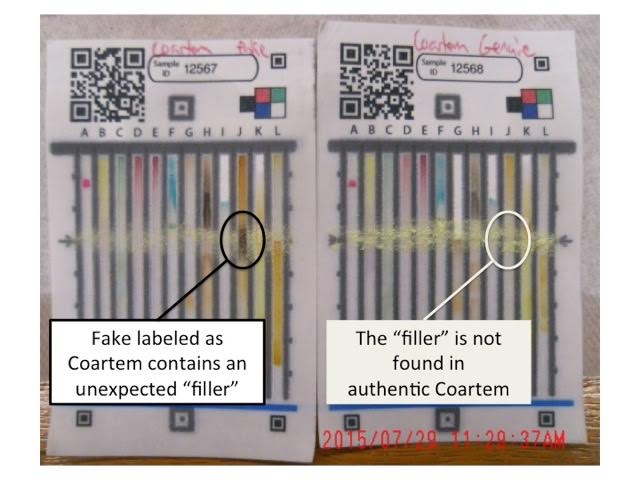}
\end{center}
   \caption{PAD images with fake and authentic versions of the drug \emph{coartem}.}
\label{fig:PADfake}
%\label{fig:onecol}
\end{figure}

Billions of people in developing countries are facing the problem of fake or substandard pharmaceuticals. Poor quality medications may contain harmful or inert ingredients that fail to treat the patient's underlying condition. Drugs which contain sub-therapeutic levels of an active ingredient contribute to development of drug-resistant pathogens. The human toll of these products is large. Falsified anti-malarials have been estimated to kill 120,000 children under the age of 5 each year \cite{AntiMalaria}. Although there are pharmacopeia methods for analysis of medications, they often cannot be applied in the developing world due to lack of technical and regulatory resources, and studies consistently show that 15 - 30\% of medications for sale in Africa, southeast Asia, and other low resource settings are substandard or fake. To address this problem Weaver et al. proposed the idea of using inexpensive (under \$1) paper analytical devices (PADs) that contain libraries of color tests to respond to the chemicals inside a pill \cite{AChem}. The PADs have 12 vertical ``lanes" in which chemical reagents are embedded. When a drug is swiped across these lanes and the PAD is dipped in water, the reagents react with the different compounds in the drug, generating characteristic color tones in the vertical lanes, usually at or above the drug swipe (see Figure \ref{fig:PADExamples}.a). The distinctive set of color tones generated for a particular drug acts like its ``fingerprint", and is considerably different from that of a counterfeit version of the same drug (Figure \ref{fig:PADfake}).

\begin{figure}[t]
\begin{center}
   \includegraphics[width=0.6\linewidth]{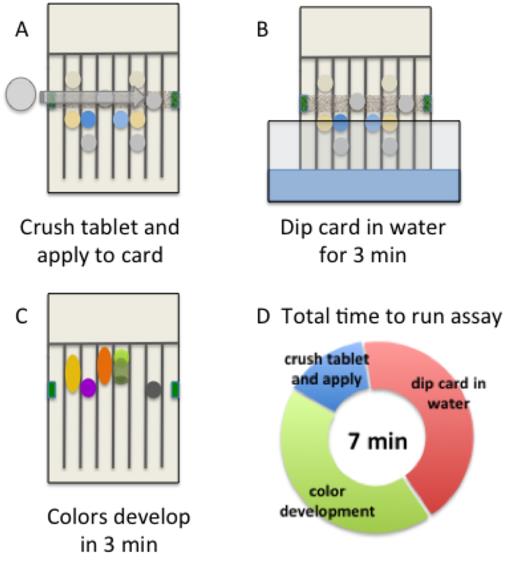}
\end{center}
   \caption{The different phases of the color testing process of a PAD.}
\label{fig:PADRunning}
%\label{fig:onecol}
\end{figure}

The different steps of the color testing process of a PAD has been illustrated in Figure \ref{fig:PADRunning}. At present, the user evaluates the results of the test by visually comparing their PAD to stored images of cards run with standard pharmaceutical samples. However, reading the card results is much more challenging than actually running the chemical tests with them. Important differences in the color tones are not always picked up by human users, and the time and effort required to train people to read the cards correctly and verify their capability is a barrier to wider scaling of the use of the cards. To tackle this, we have developed a visual recognition system which can classify a PAD image based on the pharmaceutical drug being tested.  Only 18\% of people in Africa had access to the internet in 2013 although two out of three used a mobile phone \cite{BBCArticle}, we thus decided that the image recognition system should be compatible with the cell phone network. The user should be able to submit a PAD image through her cell phone and the system should return a quality score of the tested drug back to the user. By using messaging services instead of the internet to transfer card images and the results,  this system is designed to be usable by people who live in developing countries. In this paper, we discuss the different steps of the visual recognition process of the PAD images. Different feature extraction methods and their performance in recognizing the drugs using different classifiers are also presented.

%To address this problem Weaver \etal developed inexpensive (under \$1) paper analytical devices (PAD) that contain libraries of color tests to respond to the chemicals inside a pill \cite{AChem}. The PADs have 12 vertical ``lanes" that are used to perform chemical tests in. Each of these vertical lanes can have the same or different chemical reagents embedded in them. When a drug is swiped across these lanes and the PAD is dipped in water, the reagents react with the different compounds in the drug generating characteristic color tones in the vertical lanes, usually above the drug swipe. The color tones generated for a particular drug not only differ from that of another drug for the set of reagent(s) but are considerably different from the color response of a counterfeit version of the same drug (Figure \ref{fig:PADfake}). Therefore, these color tests act like a chemical ``fingerprint" for a drug. Pharmaceutical products which contain little or no active ingredient or which include substitute ingredients give different fingerprints from authentic products. PAD image for a sample drug can be seen in Figure \ref{fig:PADExamples}.a. The different phases of the color testing process of a PAD has been illustrated in Figure \ref{fig:PADRunning}.

%At present, the human user evaluates the tests by visually comparing their PAD to images of cards run with standard samples. However, that is not a reliable method for testing the PAD images as minute differences in the color tones are not always picked up by human users. Moreover, it is extremely difficult to train human users for consistent visual evaluation in all the regions affected by counterfeit pharmaceuticals. To tackle this, we have developed a visual recognition system which can classify a PAD card image based on the pharmaceutical drug being tested. The user should be able to submit a PAD image through her cell phone and the system should return a quality score of the tested drug back to the user. By using messaging services instead of the internet to transfer card images and the results, this system should benefit people inhabiting in low resource settings. In this paper, we discuss the different steps of the visual recognition process of the PAD images. Different feature extraction methods and their performance in recognizing the drugs using different classifiers are also presented.

The remainder of this paper is structured as follows. Prior research relevant to this problem, and how our work is different from them, is discussed in section 2. In section 3 we describe the images we prepared for this work and the methods used to decide on a design choice for the PAD cards. Section 4 explains the different feature extraction methods and the classifiers used for the experiments. The experimental results are discussed and further analyzed in section 5. Section 6 concludes the paper.

\section{Related work}
Modern medicine has reached great heights in the last one or two decades and it has brought about a huge growth of pharmaceuticals in the market. However, the police, regulators, healthcare providers, and patients are often faced with the task of identifying a mystery pill or tablet. Much of the work that has been done on automatic drug identification is from the last decade. Most of the researchers have discussed automatic pill recognition protocols based on the color, shape and imprint information on the pills in \cite{PillRec1,PillRec2,PillRec3}. The work of Lee \etal in \cite{PillJain2} is worth mentioning in this regard, where they try to retrieve images of similar pills when provided with a query pill image. They extract the color histogram and the Hu moments \cite{HuFeat1} based on the shape of the pill as features. Additionally, SIFT \cite{SIFTFeat1} and MLBP \cite{MLBPFeat1} descriptors are used to encode any imprint information present on the pills. The matching accuracy for these pills is computed using both individual features and their different combinations. They report a rank-1 retrieval accuracy of about 73\%. Such pill identification systems, such as \cite{PillRecPage1,PillRecPage2,PillRecPage3}, are available online for general purpose use.

Our work is considerably different from what has been proposed in these papers as we crush the pills and swipe them on the PAD cards for classification based on their chemical composition. Thus, instead of identifying the pill based on its appearance we recognize the drug based on its chemical composition, which is revealed by the chemical ``fingerprint" it makes in reacting with the reagents embedded in the PAD cards. The main objective of our work is to detect pills whose formulations do not match the authentic version.  Fake pills can match the real ones exactly in appearance, but if the ingredients of the pill are different, the color fingerprint generated on the PAD will be considerably different from that of the real drug. An additional advantage of the PAD method is that some types of falsification can be detected even if an authentic sample of the particular brand of pill is not available. To our knowledge, there is no published work in the literature which uses vision techniques for the recognition of the PADs and our work is the first of its kind.

%Our work is considerably different from what has been proposed in these papers. Firstly, instead of identifying the pill based on its appearance we recognize the drug based on its chemical composition. The features are extracted in this case not from the pill but the chemical ``fingerprint" it makes by reacting with the reagents embedded in the PAD cards. Secondly, the main objective of our work is to distinguish real drugs from their counterfeit versions, which in many cases can be extremely difficult or impossible using the methods from these prior papers. This is because the fake drugs can match the real ones exactly in resemblance generating identical feature vectors. However, the color tones these fake drugs generate on the PAD cards are considerably different from that of the real drug. As a result, they generate better features for classification between these two sets of cards.

\section{Preparation of the PAD cards and their images}
For correct identification of different pharmaceuticals using the PAD images, the cards have to be first prepared with the best set of up to 12 reagents (one for each lane) which produces the most unique color tones for the different drugs. For our experiments, 26 pharmaceutical drug samples\footnote{The drugs are: \emph{acetaminophen}, \emph{acetylsalicylic acid}, \emph{amodiaquine}, \emph{amoxicillin}, \emph{ampicillin}, \emph{artesunate}, \emph{azithromycin}, \emph{calcium carbonate}, \emph{chloramphenicol}, \emph{chloroquine}, \emph{ciprofloxacin}, \emph{corn starch}, \emph{DI water}, \emph{diethylcarbamazine}, \emph{dried wheat starch}, \emph{ethambutol}, \emph{isoniazid}, \emph{penicillin G}, \emph{potato starch}, \emph{primaquine}, \emph{quinine}, \emph{rifampicin}, \emph{streptomycin}, \emph{sulfadoxine}, \emph{talc} and \emph{tetracycline}.} were chosen either because they have been reported in the literature as ingredients in falsified formulations (\eg corn starch, acetaminophen) or because they are in the list of WHO essential medicines \cite{WHOList}, with high sales volume in the developing world (\eg amoxicillin, isoniazid). For this 26 drug samples, we choose the best 12 from a set of 24 reagents. In order to analyze the card images, the raw unaligned image taken with a cell phone needs to be rectified and properly aligned prior to the feature extraction process. Therefore, the card preparation and image rectification are two key steps in proper drug classification. We describe the PAD image alignment process first, and then the optimal reagent selection process below.

%For correct identification of different pharmaceuticals using the PAD images, the cards have to be first prepared with the best set of up to 12 reagents (one for each lane) which produces the most unique color tones for the different drugs. For the 26 pharmaceutical drug samples\footnote{The drugs are: acetaminophen, acetylsalicylic acid, amodiaquine, amoxicillin, ampicillin, artesunate, azithromycin, calcium carbonate, chloramphenicol, chloroquine, ciprofloxacin, corn starch, DI water, diethylcarbamazine, dried wheat starch, ethambutol, isoniazid, penicillin G, potato starch, primaquine, quinine, rifampicin, streptomycin, sulfadoxine, talc and tetracycline.} used in our experiments, we choose the best 12 from a set of 24 reagents. Once the cards are chemically tested with the questioned drugs, the raw unaligned image taken with a cell phone needs to be rectified and properly aligned prior to the feature extraction process. Therefore, the card preparation and image rectification are two key steps in proper drug classification. We describe the PAD image alignment process first, and then the optimal reagent selection process below.

\subsection{PAD image rectification}
The PAD cards are photographed after the reactions between the drug and each reagent have taken place. However, the lanes in the resulting image are unaligned and don't match the locations of the original PAD artwork (Figure \ref{fig:PADExamples}.b). So, some transformation is required before the analysis can proceed. The images have varying resolution, scaling, rotation, offsets and perspective attributes to correct for. The method we use to accomplish this is described below.

\begin{figure}[t]
\begin{center}
   \includegraphics[width=1.0\linewidth]{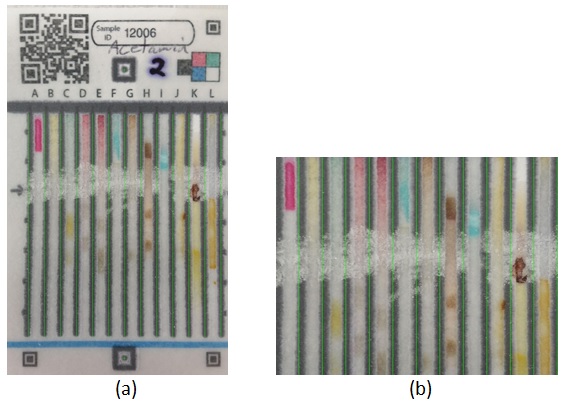}
\end{center}
   \caption{(a) The rectified version of the PAD image from Figure \ref{fig:PADExamples}, (b) the cropped out salient region from the rectified PAD image.}
\label{fig:AlignCrop}
%\label{fig:onecol}
\end{figure}

The PAD card is pre-printed with a QR code in the upper left hand corner (containing the image metadata) and three additional fiducial marks at the remaining corners. The PAD processing software uses a variant on the QR code alignment marker finding algorithm to locate the fiducial marks, together with the QR code markers, to acquire a maximum of six reference points. The resulting points are processed with the popular OpenCV \cite{OpenCV} library to do a offset, scaling, rotation and perspective correction to match the coordinates of the artwork. In addition to the ink print layer, the face of the card is printed with wax lines that are baked into the paper to separate the reagent lanes. Since the lanes can be misaligned with the original ink layer due to the printing process, two wax fiducial markers are added and a template matching scheme within OpenCV is used to find their coordinates. These coordinates are used to remove the offset and rotation of the lanes within the rectified version of the PAD image, example shown in Figure \ref{fig:AlignCrop}.a (730x1220 in size).

\subsection{Selection of the optimal set of reagents}
\begin{figure}[t]
\begin{center}
   \includegraphics[width=0.8\linewidth]{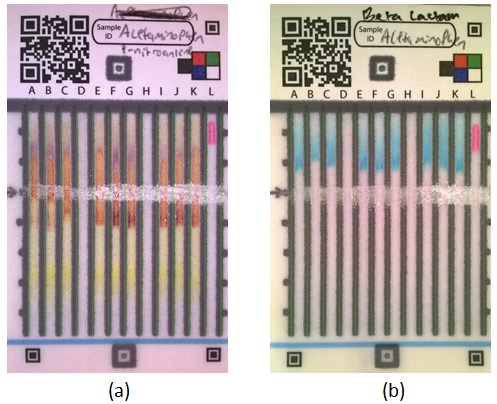}
\end{center}
   \caption{(a) Rectified PAD image of the drug acetaminophen with the reagent that detects phenol groups in the 9 lanes from A - K, (b) rectified PAD image of acetaminophen with the reagent that detects beta lactam groups in the 9 lanes from A-K. The drug concentration increases from the left to the right of image, the pink hue in lane L is from the timer agent.}
\label{fig:PAD1Reagent}
%\label{fig:onecol}
\end{figure}

24 different reagents were tested to determine which ones produced the most useful color outcomes for classification of the 26 pharmaceuticals and fillers tested in this study. We prepared a set of cards in which 9 of the 12 lanes contained a given reagent;  the lanes were arranged in groups of three with a spacer between each group (Figure \ref{fig:PAD1Reagent}). One of 26 pharmaceuticals or fillers was swiped across these lanes to give groups of light, medium and heavy concentrations of the analyte.

%Majority of pharmaceutical drugs produce a distinct color when reacting with a particular reagent, however a combination of color tests for any set of 12 reagents might not be a good fingerprint for that drug. This is due to the fact that many reagents don't react with the active ingredients of certain drugs and consequently the vertical lanes in such cases don't produce any unique reaction color. Therefore, embedding such non-reacting reagents in the card lanes only hampers the drug classification. To identify the best 12 reagents, we prepared a set of single reagent cards with the 26 drugs, with 9 out of the 12 lanes being embedded with that reagent. The drug is swiped across these lanes in light, medium and heavy concentrations. Example images of such cards can be seen in Figure \ref{fig:PAD1Reagent}.

\begin{figure}[t]
\begin{center}
   \includegraphics[width=1.0\linewidth]{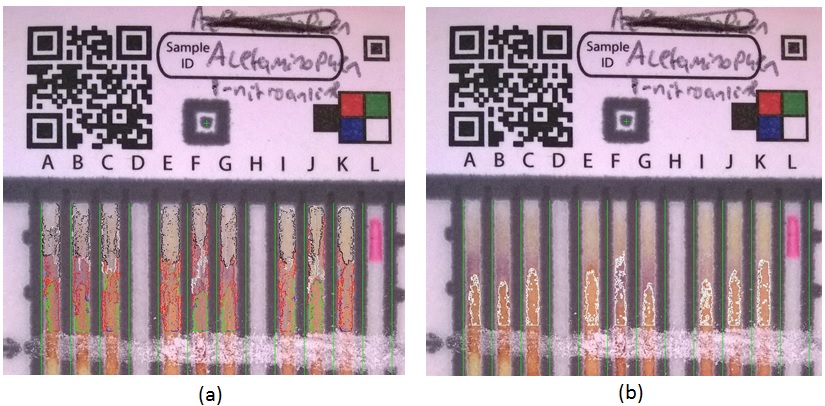}
\end{center}
   \caption{(a) The overlapping blob contours generated by the region growing algorithm in each lane, (b) the estimated actual reaction blob contour in each lane.}
\label{fig:BlobAnalysis}
%\label{fig:onecol}
\end{figure}

The cards were activated by dipping the bottom edge in water, as shown in Figure \ref{fig:PADRunning}. The different reagent/analyte combinations produce a variety of colors, and often a single lane gives multiple colors in different locations, which can be difficult for a human to interpret. Because the chemical reactions are designed to generate strong colors, we consider the color ``blobs" having the color intensity farthest from white (the background card color) to be the most meaningful color result of that lane (the reaction blob). In order to locate it on the image, we first divide the lane space above the swipe line into five equal regions. Then we run a region growing algorithm from the centroid of each region, and compute the 5 connected components from them. Since the regions are close to each other and there are at most 1 or 2 residual blobs, multiple connected components are from the same blob and they overlap, as can be seen in Figure \ref{fig:BlobAnalysis}.a.

%Although the drugs produce a blobs of a distinct color tone above the swipe line reacting with a reagent, often they produce faint (sometimes strong) residual blobs of different colors due to diffusion. The purple and faint brown blobs in the lanes in Figure \ref{fig:PAD1Reagent}.a are diffusion blobs while the dark brown blob just above the swipe line is the actual reaction blob. A maximum of one or two such extraneous blobs are produced in each lane, making it difficult for human user to predict which one is the actual reaction blob in that lane. Since the color tones grow fainter in the diffusion blobs, we consider the blob having the color intensity farthest from white (the background card color) to be the most meaningful blob. In order to locate it on the image, we first divide the lane space above the swipe line into five equal regions. Then we run a region growing algorithm from the centroid of each region, and compute the 5 connected components from them. Since the regions are close to each other and there at most 1 or 2 residual blobs, multiple connected components are from the same blob and they overlap, as can be seen in Figure \ref{fig:BlobAnalysis}.a.

We merge two such regions together if the number of overlapping common pixels between the regions is greater than a threshold. Experiments have shown the overlapping regions to merge best when this threshold is set as 0.35 times the size (\# of pixels) of the bigger region of the two. Once this is done, we end up with 2 or 3 blobs in total per lane, one of which is the actual reaction blob and the others are residual. Since the residual blobs are generally fainter in color intensity than the reaction blob, we compute the maximum difference across the R, G and B channels of the mean pixel intensity for each blob as shown in equation \eqref{eq:maxDiff}. We keep the blob with the highest value of \emph{maxDiff} per lane as it is the farthest from the background card color (white), as shown in Figure \ref{fig:BlobAnalysis}.b.

\begin{equation}\label{eq:maxDiff}
\begin{split}
\emph{\mbox{maxDiff}} = max(\mid R_{mean}-G_{mean} \mid, \mid G_{mean}-B_{mean} \mid, \\
\mid B_{mean} - R_{mean} \mid)
\end{split}
\end{equation}

9 reaction blobs are obtained in this way in the 9 lanes of the PAD image. The mean RGB intensity of these 9 blobs serve as a reaction color descriptor for a particular drug and a reagent. A database of such descriptors is built for all the possible pairs from the 26 drugs and 24 reagents. We investigate the uniqueness of each point in the database for a particular drug using the singular value decomposition (SVD) method. A 26x24 matrix (\emph{M}) is built which contains the Euclidean distance of the mean RGB intensities between the drug and each reagent. Decomposing \emph{M} by applying SVD, shown in equation \eqref{eq:SVD}, we obtain the singular matrices \emph{U} and \emph{V}, and the diagonal matrix \emph{S}.

\begin{equation}\label{eq:SVD}
M = USV^{t}
\end{equation}

%\begin{figure}[t]
%\begin{center}
%%\fbox{\rule{0pt}{2in} \rule{0.9\linewidth}{0pt}}
%   \includegraphics[width=0.8\linewidth]{Images/NormPlot3D1.jpg}
%\end{center}
%   \caption{3D bar plot showing norm values between drug sample pairs for the optimal reagent list. Some samples are missing as sample indices ranges from 0-40 while we experiment with 26 drugs.}
%\label{fig:3DBar}
%%\label{fig:onecol}
%\end{figure}

%\begin{figure}[t]
%\begin{center}
%%\fbox{\rule{0pt}{2in} \rule{0.9\linewidth}{0pt}}
%   \includegraphics[width=0.6\linewidth]{Images/salientCrop.jpg}
%\end{center}
%   \caption{The cropped out salient region from a rectified PAD image of the drug Acetaminophen with the 12 optimal reagent configuration.}
%\label{fig:salient}
%%\label{fig:onecol}
%\end{figure}

\emph{V} is a 24x24 matrix which contains information about the contribution of each reagent to the singular values of \emph{M}. Sorting the columns of \emph{V} in a descending order of corresponding singular value lists the indices of the reagents by their uniqueness for the particular test drug. They can be termed as the optimal reagents for that drug. We capture the first index from the sorted list, \ie the index of the reagent that produces the most distinctive reaction color with that drug, for all the 26 drug samples in a list. The list contains 9 different reagents, as many reagents react very distinctively with multiple drugs. Based on this list, we compute the norm between all drug sample pairs from information in matrix \emph{M} to verify the uniqueness of the results. All samples are found to have an inter-class norm of more than at least 1 intra-class standard deviation of that drug. We added 2 more reagents which are known to produce unique results from their chemistry to the list\footnote{Optimal reagents (lane A to lane L): Ni/nioxime timer, ninhydrin test for primary amines, biuret reagent, acidic cobalt thiocyanate, neutral cobalt thiocyanate, copper test for beta lactam, sodium nitroprusside, napthaquinone sulfonate, copper test for ethylenediamines, iodine test for starch, phenol test and ferric ion.}. We then prepared new 12-lane cards with a timer lane and the 11 optimal reagents, as shown in Figure \ref{fig:PADExamples}.a, and used these cards to generate color fingerprints from the 26 active pharmaceutical ingredients and excipients. The colors from the chemical reactions between the drug and the reagents are formed either at or above the swipe line, so we crop out these salient regions from the rectified PAD images (resolution of 636x490), and store them in the database, discarding the rest of each image (Figure \ref{fig:AlignCrop}.b).

%The chemical tests observed from the drug-reagent reactions are mainly concentrated above the swipe line and in some cases just below it in each lane (salient region). Therefore, the images contain a lot of redundant information. To help the feature extraction and classification process, we crop out salient regions from rectified PAD images (resolution of 636x490), and store them in the database, discarding the rest of each image (Figure \ref{fig:AlignCrop}.b).

%Based on the set of optimal reagent list, we compute the norm between all drug sample pairs from information in matrix \emph{M}. All samples are found to have an inter-class norm of more than at least 1 intra-class standard deviation of that drug. A visual representation of the sparse matrix created from this norm data can be seen in Figure \ref{fig:3DBar}.

\section{Feature extraction and classification}
Since the PAD images are not just a set of color tones but also contain characteristic blob patterns, we explored a broad set of features, ranging from low-level to high-level. Each feature used in our experiments and the classifiers used for it are described below.

{\bf L*a*b* color histogram}: The primary distinguishing factor for classifying two PAD images of two different drugs is by comparing their color fingerprints. Some of the images have reaction blobs restricted to a certain number of distinguishable colors while others span across a wide variety of color tones. We capture a 90-dimensional binning of the L*a*b* histogram as a feature that characterizes the color distribution in the images.

{\bf GIST descriptors}: We use the GIST descriptors (512 dimensions) as another low-level feature as it describes the spatial envelope of the image and provides a good representation of the visual field \cite{GISTFeat1}.

{\bf Color bank}: This is the color bank feature proposed in \cite{NIPSFeatExt}. The PAD image is first converted to a set of color names \cite{ColorFeat1,ColorFeat2} and the histogram of these names is extracted from dense overlapping patches of different sizes from the images. We learn a dictionary (of size 20) using k-means \cite{ColorFeat3} from a random sampling of the patch histograms and apply locality-constrained linear coding (LLC) \cite{ColorFeat4} to soft-encode each patch to dictionary entries. Then, max pooling is applied with a spatial pyramid \cite{ColorFeat5} on the dictionary to obtain the final feature vector (420 dimensions).

{\bf Dense SIFT}: The SIFT descriptor \cite{SIFTFeat1} captures the local patterns in the image in a scale and orientation invariant manner. We extract the SIFT descriptors from a dense grid over the images at different patch sizes. A dictionary is obtained by applying k-means (k = 256) on these descriptors and we soft encode the dictionary entries with LLC. Then the final feature vector (5376 dimensions) is obtained by applying max pooling with a spatial pyramid on the dictionary.

\begin{figure}[t]
\begin{center}
   \includegraphics[width=1.0\linewidth]{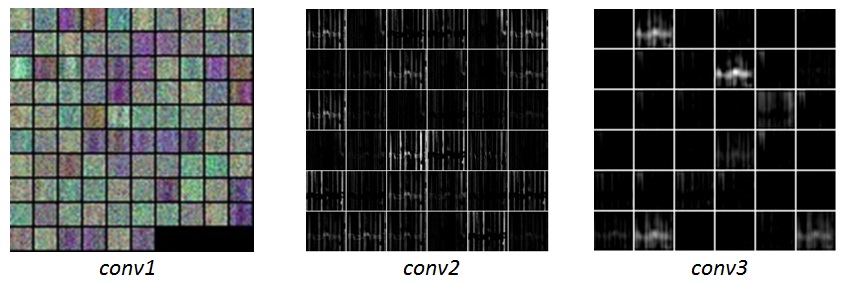}
\end{center}
   \caption{The visual representation of the filter and their responses from the 1st, 2nd and 3rd convolutional layers (\emph{conv1}, \emph{conv2} and \emph{conv3} respectively) in Caffenet.}
\label{fig:convFeat}
%\label{fig:onecol}
\end{figure}

{\bf Convolutional neural network}: The recent resurgence of deep learning models and their high performance in visual recognition tasks shows a lot of promise in this regard. Instead of hand-designed features traditionally used for visual recognition, the convolutional neural network (CNN) models extract features automatically in their intermediate layers from the input images while training the network. The fact that CNN based models have been winning the last few editions of the ImageNet Large Scale Visual Recognition Challenge (ILSVRC) \cite{ILSVRC}, which deals with recognizing objects from a million images, underpins its robustness and the reliability of its automatic feature design process. We use Caffe \cite{Caffe} for feature extraction, training and testing two popular model architectures with our data, as described below.

\begin{itemize}
\item {\bf Caffenet}: It is a clone of the network (5 convolutional + 3 pooling layers) which won the ILSVRC 2012 contest proposed by Krizhevsky \etal in \cite{AlexNet} for general object recognition. We use only the architecture as it is, except in the last fully connected layer where the number of output is changed to 26, and train the network with our PAD image data from scratch. The features, both low and high level, are automatically captured by the network in the filter maps of its intermediate layers, as shown in Figure \ref{fig:convFeat}.

\item {\bf GoogLeNet}: It is the ILSVRC 2014 winning ``very deep" 22-layer network architecture from Google, described in \cite{GoogLeNet}. We use the architecture only, with the configuration of the fully connected layers changed, and train it with the PAD images from scratch. The features are captured in the intermediate layer filters.
\end{itemize}

To train and test the features discussed above, except those from the CNN models, we use two classifiers: a simple squared-Euclidean distance based nearest neighbor (kNN, where k = 1) and a multi-class support vector machine (SVM) with a radial basis kernel. The SVM training parameters are estimated using cross-validation on the dataset. We chose the softmax regression layer used for classification in the original networks described above, for training and testing with the CNN features. Thus, we use the entire pipeline of these models for both feature extraction and classification of the PAD images.

\section{Experiments and results}
For our experiments we prepare a dataset of 30 images (cropped salient regions) per drug, making a total of 780 images for the 26 drugs. We randomly select 20 images for training for each drug and use the remaining 10 for testing, giving us a total of 520 images for training and 260 for testing. The classifiers (kNN and SVM) are trained using the features described in the previous section. The CNN models train with the features they capture from the images automatically. We set a maximum cap of 100,000 iterations for training for both the models as our dataset is much smaller compared to ImageNet \cite{ImageNet}. The network models tend to converge earlier than that, generally between the 70,000 - 80,000 iterations. The weight configuration (model state) which gives the best performance on the validation images is selected for testing. Since our data set is small in size, to handle overfitting (especially for the CNN models) on the training data, we perform training from scratch using a 3-fold cross validation approach on the available dataset and calculate the classification accuracy by averaging results from each experiment. The top-1 accuracy, averaged over the 3 folds, of each method in correctly classifying the test PAD images can be seen in Table \ref{Tab:MethodsAcc}.

\begin{table}
\caption{Average classification accuracy of PAD images}
\begin{small}
\begin{center}
\begin{tabular}{  |c | c|  }
\hline
  Method & Accuracy (\%) \\
\hline
\hline
  L*a*b* histogram with kNN  &  138/260 (53.07\%)  \\
  L*a*b* histogram with SVM  &  172/260 (66.15\%) \\
  \hline
  GIST with kNN & 216/260 (83.07\%)\\
  GIST with SVM & 230/260 (88.46\%)\\
  \hline
  Color bank with kNN & 235/260 (90.38\%)\\
  Color bank with SVM & 231/260 (88.84\%)\\
  \hline
  Dense SIFT with kNN & 200/260 (76.92\%)\\
  Dense SIFT with SVM & 233/260 (89.61\%)\\
  \hline
  Color bank + dense SIFT with kNN & 238/260 (91.53\%)\\
  Color bank + dense SIFT with SVM & 240/260 (92.30\%)\\
  \hline
  {\bf Caffenet}  &  {\bf 245/260 (94.23\%)} \\
  \hline
  GoogLeNet & 243/260 (93.46\%)\\

\hline
\end{tabular}
\label{Tab:MethodsAcc}
\end{center}
\end{small}
\end{table}

\begin{figure*}
\begin{center}
   \includegraphics[width=1.0\linewidth]{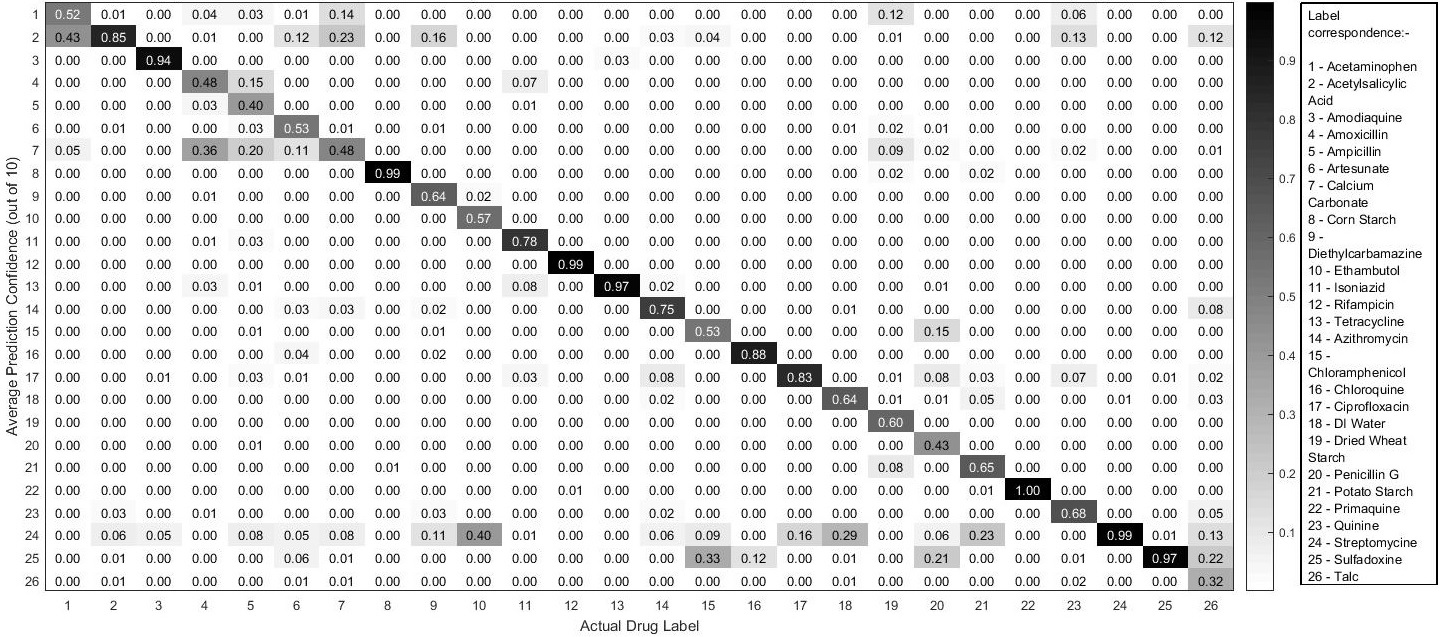}
\end{center}
   \caption{The average prediction confidence for classifying each drug class in Caffenet. The diagonal values represent the prediction confidence for that drug index.}
\label{fig:Prediction}
%\label{fig:onecol}
\end{figure*}

Clearly, some methods are better than others in classifying the cards. For example, with the color histogram alone as the feature almost half the test images are misclassified while the predictions are much more accurate when using higher level features. However, no one method can be marked as the outright winner in a statistical sense, as their confidence intervals overlap over the 3 folds. Interestingly, both the CNN models do a good job in classifying the PAD images, with Caffenet churning out the most accurate average predictions for our dataset. The average confidence for each drug class that Caffenet makes a prediction for can also be found in Figure \ref{fig:Prediction}. Although the data is limited, still the network is quite precise in making the accurate predictions in majority of the cases considering that chance prediction confidence is 0.038 (1/26). Caffenet slightly outperforming (less than 1\%) GoogLeNet is suggestive of the fact that deeper features of the latter don't contain any additional information for our small and well structured dataset.  We anticipate GoogLeNet to outperform Caffenet on a larger dataset of PAD images which have higher variance in them.

The CNN model architectures we used are originally designed for general object recognition, training with thousands of images for feature extraction. However, the images that are used for training have a high level of variance, both intra-class and inter-class. Our dataset, although much smaller in size, has images which are not so random and much more structured. That might be the reason behind the feature learning mechanisms of these architectures transferring well to a specific recognition task like ours. To explore this further we decided to run some additional experiments on Caffenet.

\begin{figure}[t]
\begin{center}
   \includegraphics[width=1.0\linewidth]{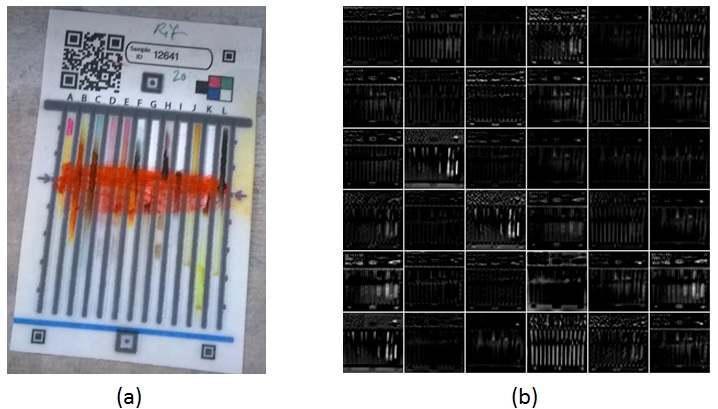}
\end{center}
   \caption{(a) A sample unrectified wild type PAD image for the drug \emph{rifampicin}, (b) the filter response generated for unrectified card by the 2nd convolutional layer (\emph{conv2}) in Caffenet.}
\label{fig:UnrectifiedFeat}
%\label{fig:onecol}
\end{figure}

\begin{table}
\caption{Average classification accuracy of non-structured PAD images with Caffenet}
\begin{small}
\begin{center}
\begin{tabular}{  |c | c|  }
\hline
  Method & Accuracy (\%) \\
\hline
\hline
  Unrectified images with Caffenet  &  233/260 (89.61\%)  \\
  \hline
  Images with random lane order with Caffenet  &  87/260 (33.46\%) \\
\hline
\end{tabular}
\label{Tab:2ndExp}
\end{center}
\end{small}
\end{table}

We predicted two factors to be contributing to this high performance on Caffenet - (1) the rectification and salient region cropping process of the raw PAD images as described in section 3 and (2) the rigid ordering of the reagents in the lanes. To check the influence of the rectification process on the performance, we built a dataset with unrectified wild type images for the same 26 drugs and optimal reagents. An example image can be seen in Figure \ref{fig:UnrectifiedFeat}.a. The images are directly used for training and testing, without any cropping of the salient region as well. We train Caffenet with 520 images as before and test with the remaining 260. The result can be found in Table \ref{Tab:2ndExp}. Although the accuracy in this case drops a little, it still does a better job in classifying the drugs than some of the other methods do with even rectified images. The filter maps, when pulled out from the network, suggest that the network locates the salient regions automatically even from the unrectified images (Figure \ref{fig:UnrectifiedFeat}.b).

\begin{figure}[t]
\begin{center}
   \includegraphics[width=1.0\linewidth]{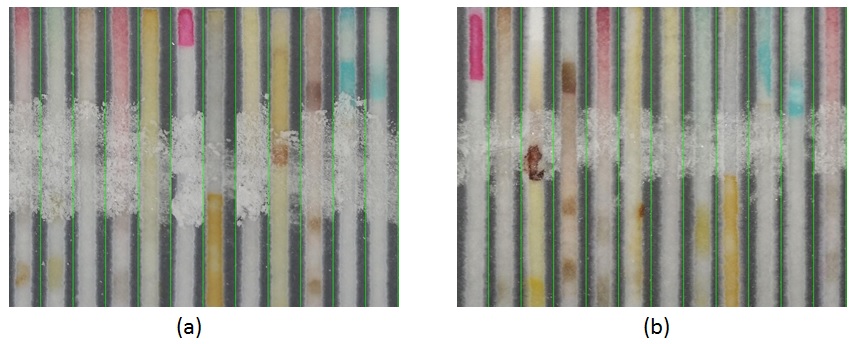}
\end{center}
   \caption{(a) A perturbed training image for the drug \emph{acetaminophen}, (b) a perturbed testing image for the same drug. It can be easily observed that the lane positions are different for the two images.}
\label{fig:Perturbed}
%\label{fig:onecol}
\end{figure}

To examine the effect of the rigid ordering of the reagents in the vertical lanes has on the classification performance, we perturb the structure of the PAD images again. We randomly re-order the lane positions in the training and testing salient crops for each drug such that the position of a reagent (lane A - lane L) in the training images is different than its position in the test images. A pair of example training and testing images can be seen in Figure \ref{fig:Perturbed}. We train Caffenet as before, with 520 perturbed images and test it with 260 images. The accuracy goes down significantly in this case, as shown in Table \ref{Tab:2ndExp}. The result clearly indicates that the reordering of the reagents hampers Caffenet's learning process as the features captured during training are much different from the test features. Thus, the network indeed looks at blob patterns in the images and not just the color tones when extracting features.

%Therefore, it can be inferred that strict ordering of the reagents is a vital factor in accurate classification of the PAD images.

%For the training images (the salient crops in the original training set) of each drug, we perturb the lane positions using a set of random indices \ie we change the position of the reagents in the card image. We re-orient the lanes in the test images using a set of random indices different from the training one. Thus, the position of a reagent (lane A - lane L) in the training images is different than its position in the test images.

\section{Conclusion}
In this paper, we have proposed the novel problem of visual classification of pharmaceutical drugs and its application in counterfeit drug detection. Unlike prior research in this area, our work doesn't use the shape, color or imprint information or the pill appearance itself for this purpose. Instead we focus on the chemical composition of the drug by observing the color tones it generates reacting with different reagents on a PAD card, like a chemical fingerprint of the drug. This facilitates the detection of fake drugs, as the change in chemical composition can easily be picked up by the chemical tests. We developed a dataset of 780 PAD cards with 26 drugs and 11 optimal reagents (and a timer agent) to conduct our experiments. The raw PAD images were aligned using a rectification process and the salient region from each image was cropped out. We extracted hand-crafted features from the salient crops and trained kNN and SVM classifiers with them. We also trained two popular CNN architectures (Caffenet and GoogLeNet), originally designed for general object recognition, with these images. Both the models performed well in accurately classifying the test card images with Caffenet giving the highest average accuracy of over 94\% on our dataset. On experimenting with unrectified and unstructured PAD images on Caffenet it was observed that the order of the different reagents in the vertical lanes of the cards is a vital factor in accurate classification of the drugs.

While our research has made a promising start, more work is needed in certain areas. To improve the classification performance we require more data \ie more physical cards to take the images from. Our collaborators in Africa have been sending in their test images (actual field data) recently and we are in the process of making a bigger database for classification with more drugs. The other option is to automate the card preparation process. Currently we manually bake, embed reagents, swipe drug and test the cards by hand in a wet lab, which is both time-consuming and tedious. We plan to automate this so that the cards are manufactured faster and in a more streamlined manner. The other possible extension of our work would be to design our own CNN architecture, instead of using one out of the box. Although Caffenet performs well in classifying the PAD images, still its originally designed for general object recognition. A network designed for a more specific object classification, like ours, should improve the performance. Lastly, we are in the process of implementing a message service based cell phone application through which any person in possession of a PAD card would be able to send us their test images. This will be the best way to get our hands on large quantities of images, including different brands of pharmaceutical products and falsified drugs, to improve the visual recognition system. If just one sample of a very poor quality medication is found using the system, additional targeted sampling, analysis, and regulatory action can be carried out to get the affected products off the market.

\vspace{0.2in}
\noindent{\bf Acknowledgements}
%\vspace{0.1in}

This work was funded by the Bill \& Melinda Gates Foundation 01818000148 and the USAID-DIV 50709. We thank Kevin Campbell and Lyuda Trokhina for their help in preparation of the PAD database. We also acknowledge Patrick Flynn, Walter Scheirer, Charles Vardeman, Aparna Bharati and John Bernhard for improving the paper with their suggestions.

\end{document}